\documentclass[graybox]{svmult}

\usepackage{type1cm}   
\usepackage{array}
\usepackage{orcidlink}

\usepackage{makeidx}         
\usepackage{graphicx}        
\usepackage{multicol}        
\usepackage[bottom]{footmisc}

\usepackage{newtxtext}       %
\usepackage{newtxmath}       

\makeindex             

\begin{document}

\title*{Evaluating Language Models on Grooming Risk Estimation Using Fuzzy Theory}
\author{Geetanjali Bihani\orcidlink{0000-0001-8352-7948}, Tatiana Ringenberg\orcidlink{0000-0002-9851-6341}, Julia Rayz\orcidlink{0000-0003-3786-2416}}
\institute{Geetanjali Bihani \at Purdue University, USA, \email{gbihani@purdue.edu}, \and Tatiana Ringenberg \at  Purdue University, USA,  \email{tringenb@purdue.edu}
\and Julia Rayz \at  Purdue University, USA,  \email{jtaylor1@purdue.edu}}
%
%
\maketitle

\abstract*{}

\abstract{Encoding implicit language presents a challenge for language models, especially in high-risk domains where maintaining high precision is important. Automated detection of online child grooming is one such critical domain, where predators manipulate victims using a combination of explicit and implicit language to convey harmful intentions. While recent studies have shown the potential of Transformer language models like SBERT for preemptive grooming detection, they primarily depend on surface-level features and approximate real victim grooming processes using vigilante and law enforcement conversations. The question of whether these features and approximations are reasonable has not been addressed thus far. In this paper, we address this gap and study whether SBERT can effectively discern varying degrees of grooming risk inherent in conversations, and evaluate its results across different participant groups. Our analysis reveals that while fine-tuning aids language models in learning to assign grooming scores, they show high variance in predictions, especially for contexts containing higher degrees of grooming risk. These errors appear in cases that 1) utilize indirect speech pathways to manipulate victims and 2) lack sexually explicit content. This finding underscores the necessity for robust modeling of indirect speech acts by language models, particularly those employed by predators.} 

\section{Introduction}
\label{sec:1}
Online child sexual grooming refers to the insidious process through which an adult establishes relationships with potential under-aged victims on digital platforms,  with the goal of eventual sexual gratification without detection \cite{wintersJeglic}. Adults grooming children use a wide range of tactics and persuasion strategies depending on factors such as their potential goals \cite{ChiuSeig, kloessQualitative, DeHart} and level of directness \cite{kloess2017, kloessQualitative}. Prior research has identified various covert and overt persuasion and manipulation strategies of groomers including gift-giving, flattery, pressure, deception, and affection, to gain the trust and confidence of their targets \cite{joleby2021offender, gam}.

Research on automated grooming detection has predominantly framed the task as a binary classification problem to categorize entire chat instances as either grooming or non-grooming \cite{preub2021, gunawan, isaza}. However, this method falls short of facilitating preventive measures. Furthermore, recent studies focusing on modeling preventive measures for grooming chats have primarily relied on training with internet sting data rather than authentic victim conversations \cite{vogt2021}, potentially leading to inaccurate generalizations. Despite findings highlighting differences in chat language across various participant groups (real victims, law enforcement officials, decoys) \cite{ring21_2}, current models neither measure nor account for such disparities in data distributions.




\section{Methodology}
\label{sec:2}
In this section, we outline our methodology for modeling grooming risk using SBERT \cite{sbert2019}. We define the task of \textit{Grooming risk-scoring} and its objectives. We then define an evaluation protocol that uses fuzzy rules to compare different degrees of grooming with transformer language model predictions. Fuzzy annotations of grooming strategies have been taken from the dataset described in \cite{ring21}.

\subsection{Degrees of Grooming Risk}

Grooming risk, rather than fitting neatly into discrete categories, exhibits gradations across multiple degrees of severity. Imagine a scenario where an array of grooming strategies is deployed within a given chat context. These strategies can encompass a spectrum of subtle to overt tactics. Each strategy contributes to the overall risk level, but the degree of influence it holds is imprecise and contextually contingent. Unlike binary categorizations, grooming risk manifests along a continuum, where various factors interplay to determine the level of vulnerability. 

To assess whether a language model can learn human perceptions of grooming risk from natural language contexts, we compare language model predictions with human perceptions of grooming risk. Specifically, we map the extent to which humans perceive grooming strategies to be present in a given chat context to the grooming risk variable. We leverage human annotations of grooming strategies as outlined in \cite{ring21}, as risk scores. For a given chat context $c$, we define the total number of observed grooming strategies $n_{s}(c)$ as the sum of individual strategy scores $s_i$, where each $s_i \in \{0, 0.5, 1\}$ represents the absence (0), partial presence (0.5), or full presence (1) of the $i^{th}$ strategy. This is shown in Equation~\ref{eq:1}. 

\begin{equation}
\label{eq:1}
n_{s}(c)=\sum_{i=1}^{N} s_{i}(c)
\end{equation}





\subsection{Task Definition}
\label{subsec:1}
We approach grooming risk-scoring as a regression task, with chat context language as the independent variable and aggregated grooming risk as the dependent variable.

\textbf{Definition 1 - Chat Context $(c)$}: A sequence of current and last $n-1$ chat messages. We fix $n=3$, for our analysis.

\textbf{Definition 2 - Grooming Risk $(r_{groom})$}: Severity of grooming behavior determined by the total number of grooming strategies present within a given chat context \textit{(c)}, as shown in Equation \ref{eq:1.1}.  
\begin{equation}
\label{eq:1.1}
r_{groom} = {n_{s}(c)} 
\end{equation}

We limit our analyses to twelve grooming strategies, including coercion, bragging, teaching, requests for images, negative comments about physique, negative comments about family, personal compliments, reverse power, asking about sexual history, checking willingness, roleplaying, and secrecy, as described in \cite{ring21}.
Each strategy is assigned a membership value of 0 (none), 0.5 (partial), or 1 (full). For a detailed description of these grooming strategies and the annotation process, refer to \cite{ring21}. Thus, chat contexts with fewer observed grooming strategies are considered lower risk, while those with more strategies present indicate higher grooming risk. 


\begin{equation}
\label{eq:2}
\mu_{risk}(c) = \varphi\left(n_{s}(c) - m\right) \text{, where }\varphi(z)=\frac{e^{-z^{2} / 2}}{\sqrt{2 \pi}}
\end{equation}
\begin{equation}
\label{eq:2.1}
\mu_{mod}(c) = \mu^{0.2}_{risk}(c) \text{ for $m=0.2$}
\end{equation}
\begin{equation}
\label{eq:2.2}
\mu_{sig}(c) = \mu_{risk}(c) \text{ for $m=1$}
\end{equation}
\begin{equation}
\label{eq:2.3}
\mu_{sev}(c) = \mu^2_{risk}(c) \text{ for $m=2$}
\end{equation}

\begin{figure}[htbp]
\centering
\includegraphics[scale=0.5]{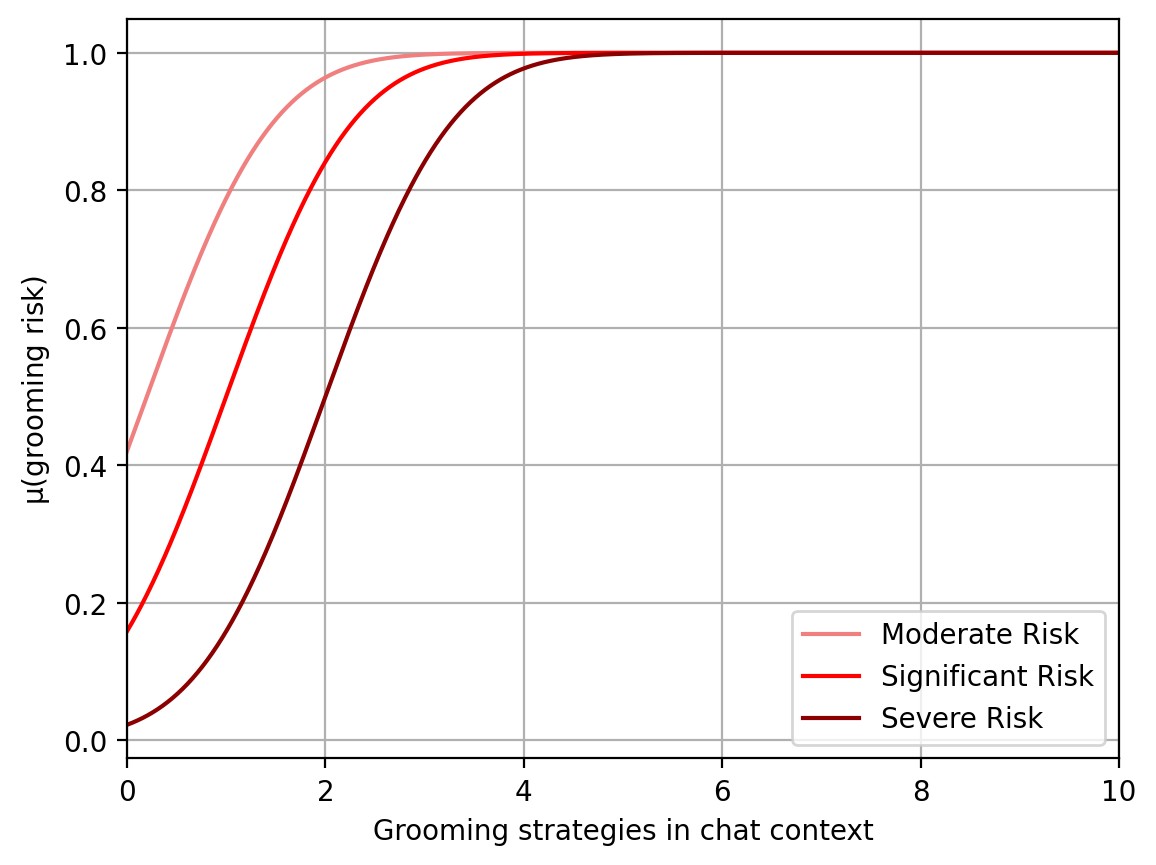}
\caption{Different degrees of grooming risk and their membership functions: Moderate [$\mu^{0.2}_{risk}(c)$], Significant [$\mu_{risk}(c)$] and Severe [$\mu^2_{risk}(c)$] risk}
\label{fig:1}       
\end{figure}

To evaluate the performance of the regression model on grooming risk prediction, we categorize these risk scores ($r_{groom}$) into three risk categories using a Gaussian membership function. This function maps continuous risk scores into degrees of membership in \textit{moderate} (low risk), \textit{significant} (medium risk), and \textit{severe} (high risk) categories. This allows each score to have fuzzy (partial) membership in multiple categories. We choose a Gaussian distribution function because it enables smooth transitions between risk levels. The discretization of the risk scores is done using Equation~\ref{eq:2}. We define different mean values ($m$) for each category's distribution function, i.e. $m=0.2$ for \textit{moderate}, $m=1$ for \textit{significant}, and $m=2$ for  \textit{severe} risk levels, as shown in Equations~\ref{eq:2.1}, \ref{eq:2.2} and \ref{eq:2.3}. These functions are visualized in Figure~\ref{fig:1}, illustrating the varying degrees of grooming risk and their corresponding membership functions. This approach aligns with the intuitive understanding that higher scores indicate greater risk while providing a more nuanced evaluation across different degrees of risk. To determine the final risk category, we apply defuzzification using a $\alpha$ cut of $0.5$, selecting the highest risk level where membership exceeds this threshold. Thus, if a chat context has memberships surpassing the $\alpha$-cut in both \textit{moderate} and \textit{severe} categories, it is classified as a \textit{severe} risk chat context.

%

\subsection{Model Studied}
\label{subsec:2}
We conducted our analysis on Sentence-BERT (SBERT) \cite{sbert2019} which was specifically designed to encode sentence embeddings. SBERT utilizes siamese networks to generate semantically meaningful representations of sentences. These representations are optimized to capture semantic similarity between sentences in a vector space, making them suitable for various downstream tasks such as semantic search and clustering.

\subsection{Fine-tuning Details}
\label{subsec:3}
We adapted SBERT to predict grooming risk scores using a linear layer on top of the final layer to output regression estimates. Fine-tuning involves adapting pre-trained models to specific downstream tasks by utilizing task-specific data. During the fine-tuning process, additional layers are added on top of the pre-trained models, and the models are trained on labeled data specific to the task at hand. Fine-tuning allows the model to acquire task-specific features and enhances the model's performance on the given task. 

For a given chat context $c$ the regression model is optimized for estimating the severity of grooming risk $r$ by minimizing the Mean Squared Error (MSE) between the predicted grooming risk $r_{pred}$ and actual grooming risk $r_{groom}$. 
The hyperparameters we used for finetuning our BERT models are listed in Table~\ref{tab:1}. 

\begin{table}
\centering
\caption{Fine-tuning Hyperparameters}
\label{tab:1}
%
%
\begin{tabular}{|c|c|}
\hline
\textbf{Hyperparameter} & \textbf{Value} \\
\hline
Optimizer & Adam \cite{kingma2014} \\
Learning Rate & $2.10^{-5}$ \\
Epochs & 5 \\
Batch size & 4 \\
\hline
\end{tabular}
\end{table}

\section{Results and Analysis}
We examined how well fine-tuned SBERT predictions align with human perceptions of grooming risk in natural language contexts. Specifically, we fine-tuned and evaluated language models separately on grooming conversations involving predators interacting with distinct groups: law enforcement officers (LEO), victims, and decoys. This analysis is prompted by prior research highlighting variations in language usage across different groups in grooming behaviors \cite{ring21}. Additionally, automated models of grooming classification and detection completely overlook these differences. This underscores the importance of demonstrating that language models trained on data distributions diverging from patterns observed in real-victim grooming conversations may lack reliability if deployed without due consideration. 

\subsection{Prediction Accuracy across Degrees of Grooming Risk}
\begin{table}[!t]
\caption{MSE reported on different groups (High values are in bold font)}
\centering
\label{tab:3}       
%
%
\begin{tabular}
{p{0.2\textwidth}p{0.2\textwidth}p{0.2\textwidth}p{0.2\textwidth}}
\hline\noalign{\smallskip}
Grooming Risk  & LEO  & Victim  & Decoy  \\
\noalign{\smallskip}\hline\noalign{\smallskip}
Moderate &  1.881 &  1.282 & 1.188 \\
Significant &  1.211 & 1.360 & 1.748 \\
Severe & \textbf{6.589} & \textbf{9.096} & \textbf{7.762} \\
\noalign{\smallskip}\hline\noalign{\smallskip}
Overall & 3.552 & 3.425 & 3.217 \\
\noalign{\smallskip}\hline\noalign{\smallskip}
\end{tabular}
\end{table}

We report Mean Squared Error on predictions across different degrees of grooming risks in Table~\ref{tab:3}. While the overall MSE remains consistent across all three groups, discernible differences in error rates are evident. Our findings show notably higher error rates for message contexts associated with \textit{severe} grooming risk, showing that models perform worse in high-risk scenarios within the context of grooming risk estimation. Moreover, a model finetuned on real-victim chats performs worse than those trained on LEO and Decoy chats. This finding highlights the differences in chat languages across groups and questions the inherent assumption made by the NLP community regarding training and finetuning automated grooming risk estimation frameworks using unrepresentative Decoy and LEO chats.

To understand the source of higher errors in \textit{severe} grooming risk contexts, we did a qualitative examination of these erroneous predictions. Our findings show that message contexts with \textit{severe} grooming risks do not necessarily utilize explicitly sexual and predatory language. Prior works on cyberforensic analyses of grooming conversations mention that predators employ direct and indirect communication pathways in grooming conversations \cite{kloess2017}. We find that in many such cases, seemingly innocuous but predatory texts remain undetected by language models, causing the error rates to increase. This limitation of fine-tuned language models arises from their tendency to learn surface-level associations, rather than learning associations within the underlying context. Prior work on measuring uncertainty in grooming language used in predatory settings across different groups \cite{ring21_2}, provides evidence of the variability in the uncertainty language used within different grooming stages. 

\subsection{Risk Distribution across Different Groups}

%
\begin{figure}[htbp]
\centering
\includegraphics[scale=.35]{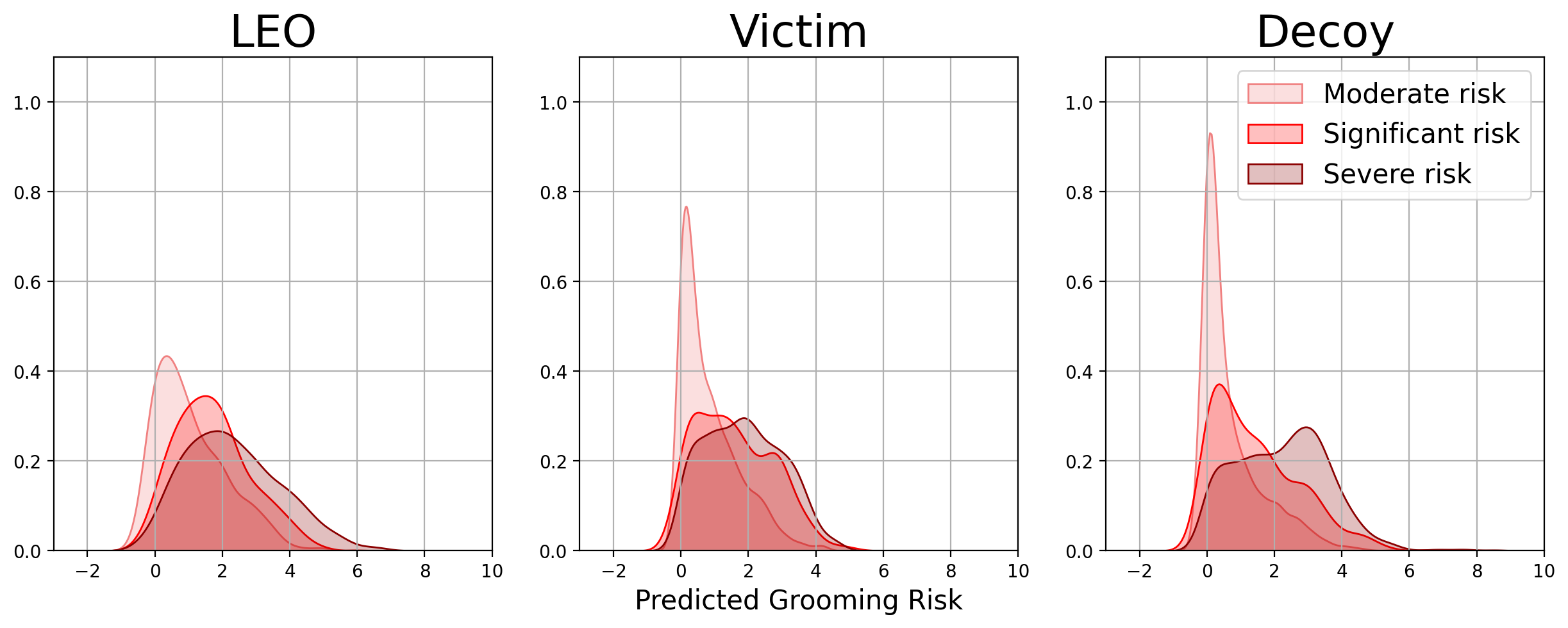}
\caption{Distribution of predicted grooming risk $(r_{pred})$ over fuzzy degrees of actual grooming risk $(r_{groom})$ varies across different groups (LEO, Victim and Decoy) .}
\label{fig:2}       
\end{figure}

We conducted an analysis of model predictions across groups, illustrating the distribution of predicted grooming risk scores ($r_{pred}$) across varying degrees of actual grooming risk, as depicted in Figure~\ref{fig:2}. Our investigation can be summarized through the following insights: while language model predictions discern between samples categorized as \textit{moderate} and \textit{severe} in terms of grooming risk, they struggle to make the same distinction between \textit{moderate} and \textit{significant} risk samples. Additionally, the distribution of predicted risk underscores the uncertainty inherent in these estimations, as evidenced by the variance across the distributions. With the exception of \textit{moderate} risk samples in Victim and Decoy conversations, our predictions exhibit substantial variance. These findings underscore the complexity involved in developing robust estimators for grooming risk within natural language contexts.

\section{Discussion and Conclusion}
This paper investigates whether SBERT can effectively discern varying degrees of grooming risk inherent in predatory conversations. We fine-tune and evaluate a language model regressor on predatory conversations across different participant groups. Our analysis highlights that while fine-tuning aids language models in learning to assign grooming scores, they exhibit high variance in predictions, particularly in contexts with higher degrees of grooming risk. This discrepancy is tied to cases where surface form text does not contain explicit identifiers of grooming, but rather utilizes indirect speech pathways to manipulate victims. In such cases, fine-tuning sentence embedding models does not help the model learn nuanced tasks. The sole reliance on word form and incentivizing training loss lead to learning shortcuts while ignoring nuance \cite{bihanirayz2024}. Even with the integration of long-range context, the task of encoding intricate lexical semantic phenomena to enhance natural language understanding continues to be a challenge \cite{bihanirayz21, vulic2021}. This finding underscores the need for robust modeling of indirect speech acts employed in grooming contexts by language models.

\begin{acknowledgement}
This work is supported by the DoJ grants 15PJDP-21-GK-03269-MECP and 15PJDP-22-GK-03107-MECP. 
\end{acknowledgement}

%
%
\bibliographystyle{plain}
\bibliography{references}
%

%
\end{document}